\documentclass[conference]{IEEEtran}

\ifCLASSINFOpdf
  \usepackage[pdftex]{graphicx}
\else
  \usepackage[dvips]{graphicx}
\fi

\usepackage[cmex10]{amsmath}
\begin{document}
\title{A Minimal Architecture for General Cognition}


% \author{\IEEEauthorblockN{Michael S. Gashler and Zachariah Kindle}
% \IEEEauthorblockA{Department of Computer Science and Computer Engineering\\
% University of Arkansas\\
% Fayetteville, Arkansas 72701\\
% Email: mgashler@uark.edu, zkindle@email.uark.edu}
% }

\author{
\IEEEauthorblockN{Michael S. Gashler}
\IEEEauthorblockA{Department of Computer Science\\and Computer Engineering\\
University of Arkansas\\
Fayetteville, Arkansas 72701, USA\\
mgashler@uark.edu}
\and
\IEEEauthorblockN{Zachariah Kindle}
\IEEEauthorblockA{Department of Computer Science\\and Computer Engineering\\
University of Arkansas\\
Fayetteville, Arkansas 72701, USA\\
zkindle@email.uark.edu}
\and
\IEEEauthorblockN{Michael R. Smith}
\IEEEauthorblockA{Department of Computer Science\\
Brigham Young University\\
Provo, Utah 84602, USA\\
msmith@axon.cs.byu.edu}
}

\maketitle

\begin{abstract}
A minimalistic cognitive architecture called MANIC is presented.
The MANIC architecture requires only three function approximating models, and one state machine.
Even with so few major components, it is theoretically sufficient to achieve functional equivalence
with all other cognitive architectures, and can be practically trained.
Instead of seeking to trasfer architectural inspiration from biology into artificial intelligence,
MANIC seeks to minimize novelty and follow the most well-established constructs
that have evolved within various sub-fields of data science.
From this perspective, MANIC offers an alternate approach to a long-standing objective
of artificial intelligence. This paper provides a theoretical analysis of the MANIC architecture.
\end{abstract}

\IEEEpeerreviewmaketitle

\section{Introduction}\label{sec_introduction}

When artificial intelligence was first discussed in the 1950's as a research project at Dartmouth \cite{McCarthy1955}, many of the early pioneers in AI optimistically looked forward to creating intelligent agents.
In 1965 Herbert Simon predicted that ``machine will be capable, within twenty years, of doing any work man can do'' \cite{Simon1965}.
In 1967, Marvin Minsky agreed, writing ``within a generation \dots the problem of creating 'artificial intelligence' will be substantially solved'' \cite{Minsky1967}.
Since that time, artificial general intelligence has not yet been achieved as predicted.
Artificial intelligence has been successfully applied in several specific applications such as natural language processing and computer vision.
However, these domains are now subfields and sometimes even divided into sub-subfields where there is often little communication \cite{McCorduck2004}.

Identifying the minimal architecture necessary for a particular task is an important
step for focusing subsequent efforts to implement practical solutions. For example, the Turing machine,
which defined the minimal architecture necessary for general computation,
served as the basis for subsequent implementations of general purpose computers.
In this paper, we present a Minimal Architecture Necessary for Intelligent Cognition (\emph{MANIC}).
We show that MANIC simultaneously achieves theoretical sufficiency for general cognition while
being practical to train.
Additionally, we identify a few interesting parallels that may be analogous with human cognition.

In biological evolution, the fittest creatures are more likely to survive.
Analogously, in computational science, the most effective algorithms are more likely to be used.
It is natural to suppose, therefore, that the constructs that evolve within the computational sciences
might begin, to some extent, to mirror those that have already evolved in biological systems.
Thus, inspiration for an effective cognitive architure need not necessarily originate from biology,
but may also come from the structures that self-organize within the fields of machine learning and artificial intelligence.
Our ideas are not rooted in the observations of human intelligence, but in seeking a simplistic architecture that could
plausibly explain high-level human-like cognitive functionality.

In this context, we propose that high-level cognitive functionality can be explained with three
primary capabilities: 1- environmental perception, 2- planning, and 3- sentience.
For this discussion, we define \textbf{Environmental perception} as an understanding of the agent's environment and situation.
An agent perceives its environment if it models the environment to a sufficient extent that it can accurately
anticipate the consequences of candidate actions.
We define \textbf{Planning} to refer to an ability to choose actions that lead to desirable outcomes,
particularly under conditions that differ from those previously encountered.
We define \textbf{Sentience} to refer to an awareness of feelings that summarize an agent's condition, coupled with a desire to respond to them.
We carefully show that MANIC satisfies criteria 1 and 2. Criterion 3 contains aspects that are not yet well-defined,
but we also propose a plausible theory for sentience to which MANIC can achieve functional equivalence.

\begin{figure}[!tb]
	\begin{center}
		\includegraphics[width=3.5in]{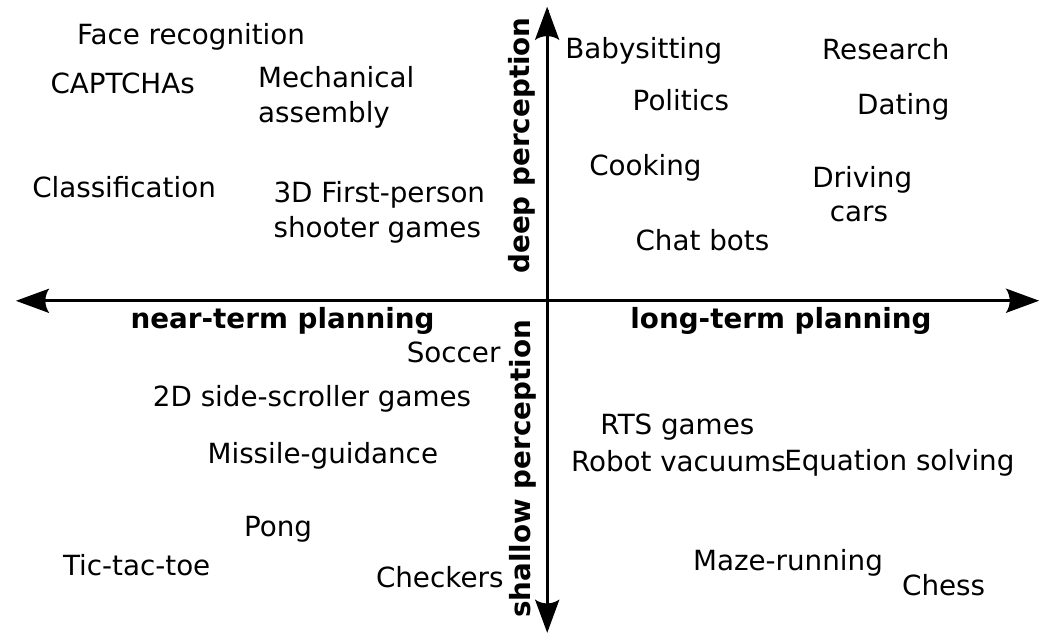}
		\caption{A subjective plot of several cognitive challenges.
			Advances in machine learning, especially with deep artifical neural networks, have solved many problems that require deep perception (top-left quadrant).
			Advances in artificial intelligence have solved many problems that require long-term planning (bottom-right quadrant).
			The unsolved challenges (mostly in the top-right quadrant) require a combination of both deep perception
			and long-term planning. Hence, the necessity of cognitive architectures, which combine advances in both sub-disciplines
			to address problems that require a combination of cognitive abilities.}
		\label{fig_problems}
	\end{center}
\end{figure}

Figure~\ref{fig_problems} shows a subjective plot of several representative cognitive challenges. This plot attempts to rank these
challenges according to our first two capabilites: perception and planning. Due to the subjective
nature of these evaluations exact positions on this chart may be debated, but it is significant to note that
the challenges typically considered to be most representative of human cognition are those that simultaneously require capabilities
in both perception and planning. Problems requiring only one of these two abilities have been largely solved in the respective
sub-discplines of machine learning and artificial intelligence, in some cases even exceeding human capabilities.
It follows that human-like cognition requires an integration of the recent advances in both machine learning and artificial
intelligence. MANIC seeks to identify a natural integration of components developed in these respective fields.

This document is laid out as follows:
Section~\ref{sec_architecture} describes the MANIC cognitive architecture.
Section~\ref{sec_sufficiency} shows that MANIC is theoretically sufficient to accomplish general cognition.
Section~\ref{sec_training} describes how the MANIC cognitive architecture can be practically trained using existing methods.
Section~\ref{sec_sentience} discusses a plausible theory for sentience, and describes how the MANIC architecture can achieve functional equivalence with it.
Finally, Section~\ref{sec_conclusion} concludes by summarizing the contributions of this paper.

\section{Architecture}\label{sec_architecture}

An implementation of MANIC can be downloaded from https://github.com/mikegashler/manic.

A cognitive architecture describes a type of software agent. It operates in a world
that may be either physical or virtual. It observes its world through a set of percepts, and
operates on its world through a set of actions. Consistent with other simple artificial
architectures, MANIC queries its percepts at regular time intervals, $t = 0, 1, 2, \ldots$, to
receive corresponding vectors of observed values, $\mathbf{x}_0, \mathbf{x}_1, \mathbf{x}_2, \ldots$.
It chooses action vectors to perform at each time, $\mathbf{u}_0, \mathbf{u}_1, \mathbf{u}_2, \ldots$.
For our discussion here, MANIC assumes that its percepts are implemented with a camera (or renderer in the case of a virtual
world), such that each $\mathbf{x}_t$ is a visual digital image.
We use vision because it is well understood in the context of demonstrating cognitive abilities.
Other percepts, such as a microphone, could also be used to augment the agent's observations.

\begin{figure*}[!tb]
	\begin{center}
		\includegraphics[width=5.8in]{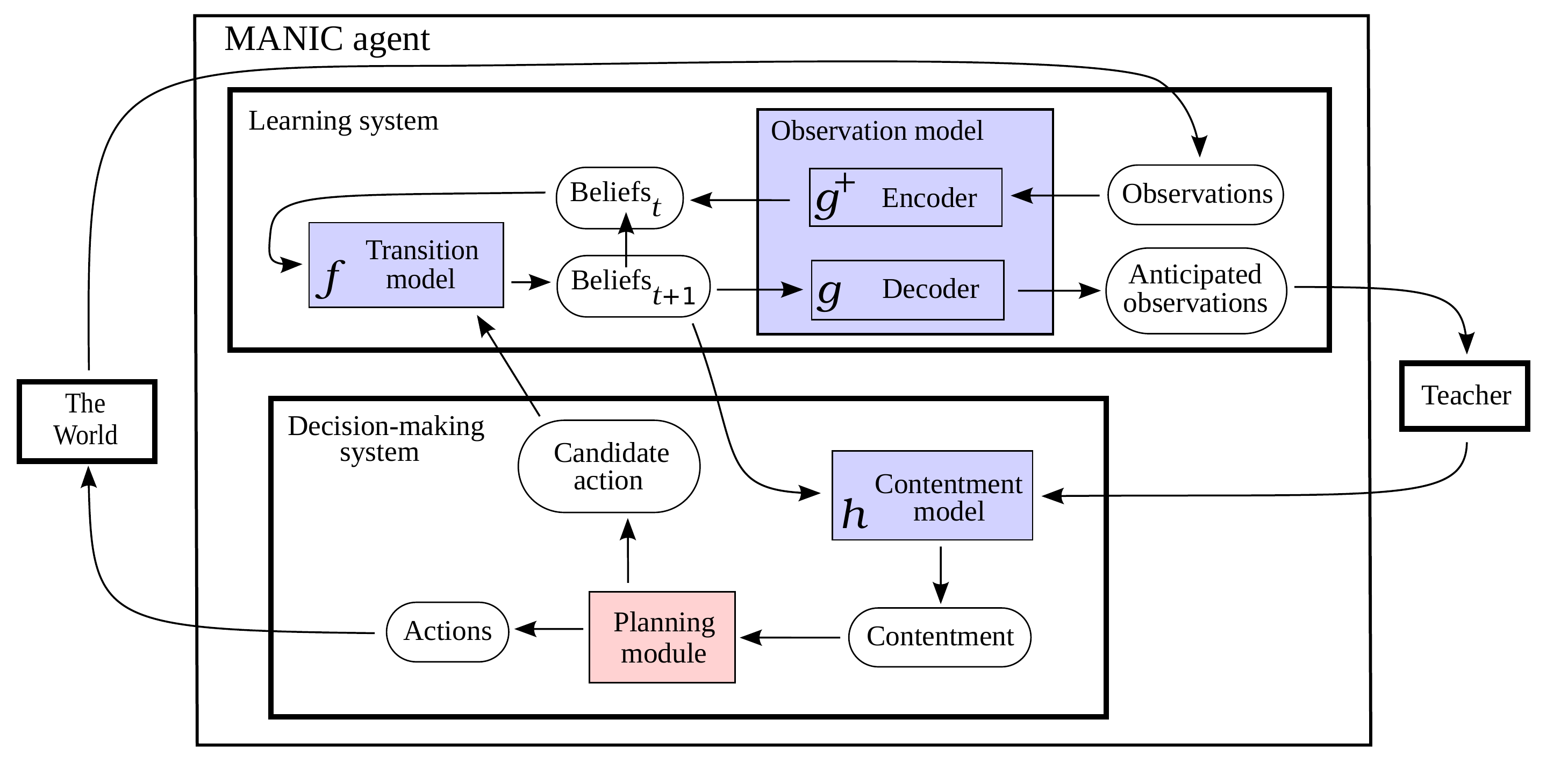}
		\caption{A high-level diagram of MANIC, which consists of just 3 function approximating models
				(blue rectangles), one of which is bi-directional, and 1 state machine (red rectangle).
				Vectors are shown in rounded bubbles.
				At the highest level, MANIC divides its artificial brain in accordance with functional
				divisions that have evolved between the sub-disciplines of machine
				learning and artificial intelligence.}
		\label{fig_agent}
	\end{center}
\end{figure*}

At a high level, the MANIC divides into two systems, which we call the \emph{learning system} and the \emph{decision-making system}.
The learning system draws its architecture from constructs that are studied predominantly in the field of machine learning, and the decision-making
system draws its architecture from the constructs that are studied predominantly in the partly overlapping field of artificial intelligence.
The agent's percepts feed into the learning system, providing the source of data necessary for learning, and the agent's actions derive from the decision-making system, which
determines the action vectors for the agent to perform.
A diagram of the MANIC architecture is given in Figure~\ref{fig_agent}.

\subsection{Learning system} The purpose of the learning system is to learn from past experiences.
A large portion of the research effort in the field of machine learning thus
far has focused on static systems. These are systems where the outputs (or labels) depend only on the
current inputs (or features) and some component of unobservable noise. With a software agent, however,
it is necessary for the agent
to model its world as a dynamical system. (A dynamical system refers to one that changes over time,
and should not be confused with a dynamic system, which is a different concept.)

Although many approaches exist for modeling dynamical systems in machine learning, nearly all of them
either explicitly or implicitly divide the problem into three constituent components:
a \emph{transition model}, $f$, a \emph{belief vector}, $\mathbf{v}_t$, and an \emph{observation model}, $g$.
The belief vector is an internal representation of the state of the system at any given time. In
the context of a cognitive architecture, we refer to it as ``belief" rather than ``state" to
emphasize that it is an intrinsic representation that is unlikely to completely represent the
dynamical system, which in this case is the agent's entire world, that it attempts to model.
The transition model maps from current beliefs to the beliefs in the next time step,
$\mathbf{v}_{t+1}=f(\mathbf{v}_t)$. It is implemented using some function approximating regression model,
such as a feed-forward multilayer perceptron. The observation model is a bi-directional mapping between
anticipated beliefs and anticipated observations, $\hat{\mathbf{x}}_t=g(\mathbf{v}_t)$, and
$\mathbf{v}_t=g^+(\mathbf{x}_t)$, where $g^+$ approximates the inverse of $g$.

When trained, this learning system enables the agent to anticipate observations
into the arbitrarily distant future (with decaying accuracy) by beginning with the current beliefs, $\mathbf{v}_t$,
then repeatedly applying the transition function to estimate the beliefs at some future time
step, $\mathbf{v}_{t+i}$. The future beliefs may be passed through the observation model
to anticipate what the agent expects to observe at that time step. Because the agent is unlikely
to ever successfully model its complex world with perfect precision, the anticipated observations
may differ somewhat from the actual observations that occur when the future time step arrives.
This difference provides a useful error signal, $e=\mathbf{x}_t-\hat{\mathbf{x}}_t$, which
can be utilized to refine the learning system over time. In other words, the agent knows its learning
system is well trained when the error signal converges toward a steady stream of values close to zero.

This learning system represents the common intersection among a wide diversity
of models for dynamical systems. For example, an Elman network \cite{gao1996modified}
is a well-established recurrent neural network suitable for modeling dynamical systems. It is typically
considered as a single network, but can be easily segmented into transition and observation components,
and its internal activations may be termed a belief vector. A NARMAX model, which is commonly used
in system identification more explicitly segments into these three components \cite{chen1989representations}.
A Jordan network is a recurrent neural network that lacks an observation model component, but it may still be
considered to be a degenerate case of an Elman network that uses the identity function for its observation
component.
%Models that predict dynamics purely through nonlinear extrapolation, such as Fourier neural
%networks \cite{silvescu1999fourier,gashler2014fourier} operate with no transition model or belief vector, but equivalent representations
%can be shown to exist that do utilize these components.
Many other approaches, including the extended
Kalman filter \cite{haykin2001kalman} and LSTM networks \cite{hochreiter1997long} naturally fit within this learning system
design.

\subsubsection{Transition model} Although the learning system as a whole is a recurrent model, the transition
model may be implemented as a simple feed-forward model, because it only needs to predict $\mathbf{v}_{t+1}$ from $\langle \mathbf{v}_t, \mathbf{u}_t \rangle$.

\subsubsection{Observation model} The purpose of the observation model is to map between beliefs and observations.
It is a bi-directional mapping between states and observations.
Of all the components in MANIC, the observation model may arguably be the most critical for
the overall success of the agent in achieving proficiency in general settings. 
If $g$ and $g^+$
are well-implemented, then the agent's internal representation of beliefs will reflect a rich
understanding of its world. This is one aspect of cognition where humans have traditionally
excelled over machines, as evidenced by our innate ability to recognize and recall images so
proficiently \cite{pinto2008real}. By contrast, machines have long been able to navigate complex
decision chains with greater effectiveness than humans. For example, machines are unbeatable
at checkers, and they can consistently trounce most humans at chess. % todo: cite
Even decision-making processes that only utilize short-term planning may appear to exhibit much
intelligence if they are based on a rich understanding of the situation.

\begin{figure*}[!tb]
	\begin{center}
		\includegraphics[width=5.0in]{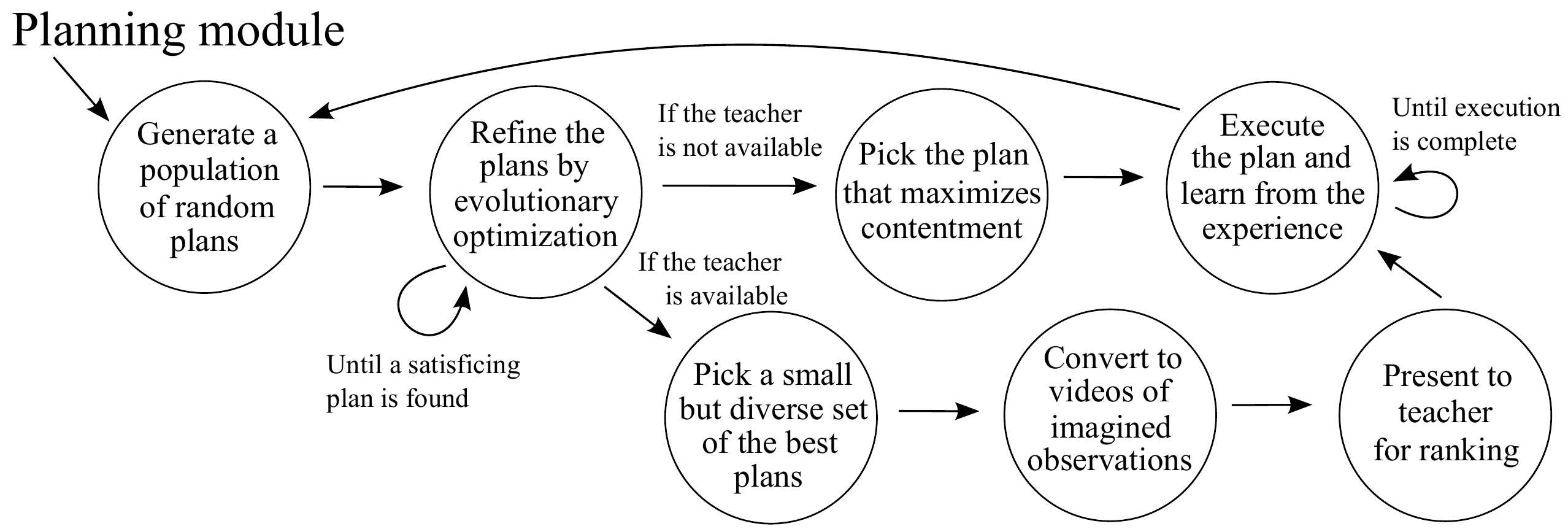}
		\caption{The planning module is state machine that makes plans by using the 3 function-approximating
			models to anticipate the consequences and desirability of candidate action sequences.}
		\label{fig_planning}
	\end{center}
\end{figure*}

Since about 2006, a relatively new research community has formed within the field of
artificial neural networks to study ``deep learning" architectures \cite{hinton2006fast,oh2004gpu}.
These deep learning architectures have demonstrated significant advances in
ability for mapping graphical images to hierarchical feature representations. For example, in
2009, Lee and others showed that deep networks decompose images into constituent parts, much
as humans understand the same images \cite{lee2009convolutional}. In 2012, Krizhevsky demonstrated unprecidented
accuracy at visual recognition tasks using a deep learning architecture trained on the ImageNet
dataset \cite{krizhevsky2012imagenet}, and many other developments in this community have eclipsed
other approaches for image recognition. % todo: cite several
Since MANIC assumes that observations are given in the form of digital images, and the task of the observation model
is to map from these observations to meaningful internal representations,
the developments of this community are ideally suited to provide the best implementation for the observation model.
The encoding component of the observation model, $g^+$, may be implemented as a deep convolutional
neural network, which is known to be particularly effective for encoding
images \cite{lawrence1997face,lecun2010convolutional,ciresan2011committee}.
The decoding component, $g$, may be implemented as a classic fully-connected deep neural network.
Instead of predicting an entire observation image, it models the image as a continuous
function. That is, it predicts all the color channel values for only a single pixel, but additional inputs
are added to specify the pixel of interest. Such an architecture can be shown to implement
the decoding counterpart of a deep convolutional neural network.

If an absolutely minimal architecture is desired, the encoder, $g^+$, may be omitted.
Rather than calculating beliefs from the current observations, beliefs may be refined to make
the anticipated observations match the actual observations.
This inference-based approach has the advantage of remembering believed state even when it is not
visible in the current observations.
Hence, the encoder is only used in the rare cases when the agent is activated in a new context,
but repeated applications of inference with the decoder would likely accomplish the same objective.

\subsection{Decision-making system} The decision making system produces plans to accomplish tasks that are anticipated to maximize the system's contentment.
Some clear patterns have also emerged among the many decision-making
systems developed for artificial intelligence. Nearly all of them divide the problem into some form of model
for estimating the utility of various possible states, and some method for planning actions that are
expected to lead to desirable states. Let $h(\mathbf{v}_t)$ be the utility that the agent believes to be
associated with being in state $\mathbf{v}_t$. Let $\mathbf{P}$ be a pool of candidate plans for maximizing
$h$ that the agent considers, where each $\mathbf{p}_i\in\mathbf{P}$ is a sequence
of action vectors, $\mathbf{p}_i=\langle \mathbf{u}_t, \mathbf{u}_{t+1}, \mathbf{u}_{t+2}, \cdots \rangle$.
At each time step, the agent performs some amount of refinement to its pool of plans, selects the one that
yields the biggest expected utility, and chooses to perform the first action in that plan.

At this high level, the model is designed to be sufficiently general to encapsulate most decision-making processes.
For example, those that maintain only one plan, or look only one time-step ahead, may be considered to be
degenerate cases of this model with a small pool or short plans. Processes that do not refine their plans
after each step may implement the regular refinement as an empty operation. Implementation details of this nature
are more properly defined in the lower-level components.

\subsubsection{Contentment model} Because MANIC is designed to be a long-living agent, we think of utility as being
maintained over time, rather only being maximized in carrying out
a single plan. Thus, we refer to $h$ as the \emph{contentment model}. We assume that any tasks the agent
is intended to accomplish are achieved through the agent's constant efforts to preserve a state of homeostasis.
For example, a robotic vacuum would be most ``content" when the floor has been recently cleaned. In order to maintain
this condition of contentment, however, it may need to temporarily stop cleaning in order empty its bag of dust,
recharge its batteries, or avoid interfering with the owner. The contentment model, $h$, is trained to capture
all of the factors that motivate the agent.

% candidate for removal
The contentment model is the only component of MANIC that benefits from a human teacher.
Because the models in the learning system learn from unlabeled observations,
direction from a teacher can be focused toward this one model.
Competitive approaches can also be used to train the contentment model without a human teacher,
as described in Section~\ref{sec_training}.

\subsubsection{Planning module} The planning module is a simple state machine that utilizes the three models to anticipate
the outcome of possible plans, and evaluate their utility.
A flow chart for this module is given in Figure~\ref{fig_planning}.
Much work has been done in the field of artificial intelligence to develop systems that 
guarantee convergence toward optimal plans. % todo: cite
Unfortunately, these systems typically offer limited scalability with respect to dimensionality \cite{bellman:curseofdimensionality}.
Early attempts at artificial intelligence found that it is much
easier to exceed human capabilities for planning than it is to exceed human capabilities for recognition \cite{pinto2008real}.
This implies that humans utilize rich internal representations of their beliefs to compensate for their
relative inability to plan very far ahead. To approach human cognition, therefore, it is necessary to
prioritize the scalability of the belief-space over the optimality of the planning.
A good compromise with many desirable properties is found in genetic algorithms that optimize by simulated evolution.
These methods maintain a population of candidate plans,
can benefit from prior planning, are easily parallelized, and scale very well into high-dimensional belief spaces.
%Unfortunately, they depend on many heuristics and do not guarantee optimal plans, but humans are not free of of such limitations either.

\section{Sufficiency}\label{sec_sufficiency}

Occam's razor suggests that additional complexity should not be added to a model without necessity.
In the case of cognitive architectures, additional complexity should probably give the architecture
additional cognitive capabilities, or else its necessity should be challenged.
In this section, we evaluate MANIC against Occam's razor, and contrast it with two even simpler
models to highlight its desirable properties.

\begin{figure}[!tb]
	\begin{center}
		\includegraphics[width=1.7in]{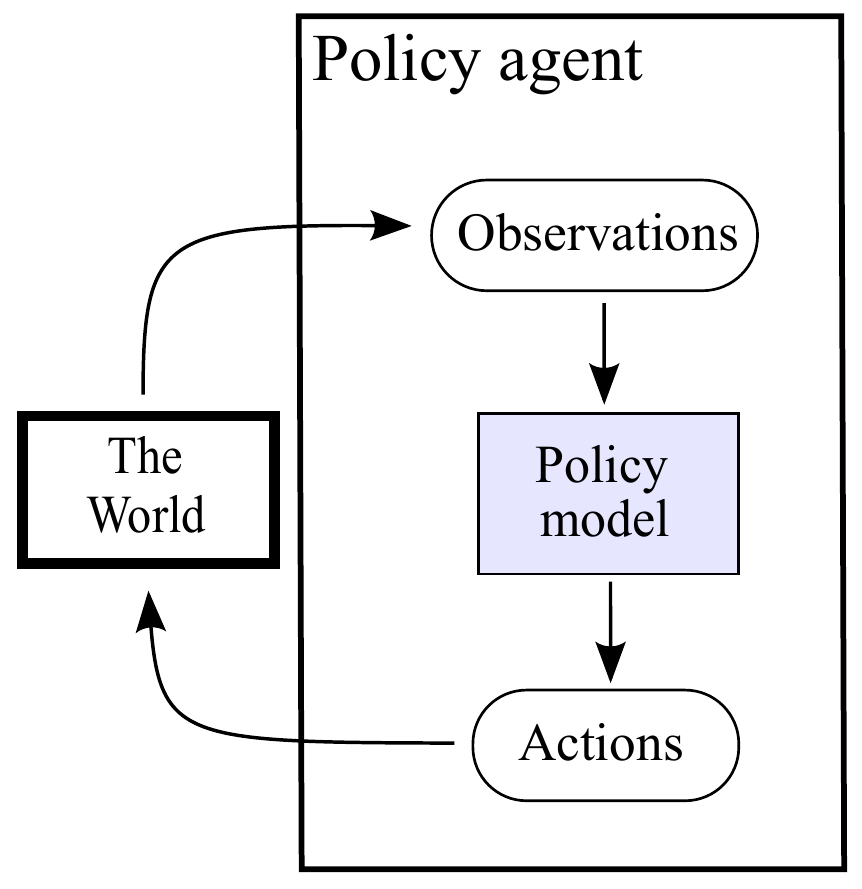}
		\caption{A simple architecture that is not sufficient for general cognition.}
		\label{fig_policy}
	\end{center}
\end{figure}

A diagram of a simple \emph{policy agent} is given in Figure~\ref{fig_policy}.
This agent uses a function approximating model to map from the current observations to actions.
(It differs from a reflex agent \cite{russell1995modern} in that its model is not hard coded for a particular problem,
but may be trained to approximate solutions to new problems that it encounters.)
The capabilities of this architecture are maximized when it is implemented with a model that
is known to be capable of approximating arbitrary functions,
such as a feedforward artificial multilayer perceptron with at least one hidden
layer \cite{cybenko1989ann_universal_function_approximators,cybenko:ann_arbitrary_func_approx}.
However, no matter how powerful its one model may be, there are many problems this architecture
cannot solve that more capable architectures can solve.
For example, if observations are not sufficient to uniqely identify state, an architecture with memory
could do better.
Therefore, we say that this policy agent is not sufficient to implement general cognition.

\begin{figure}[!tb]
	\begin{center}
		\includegraphics[width=2.8in]{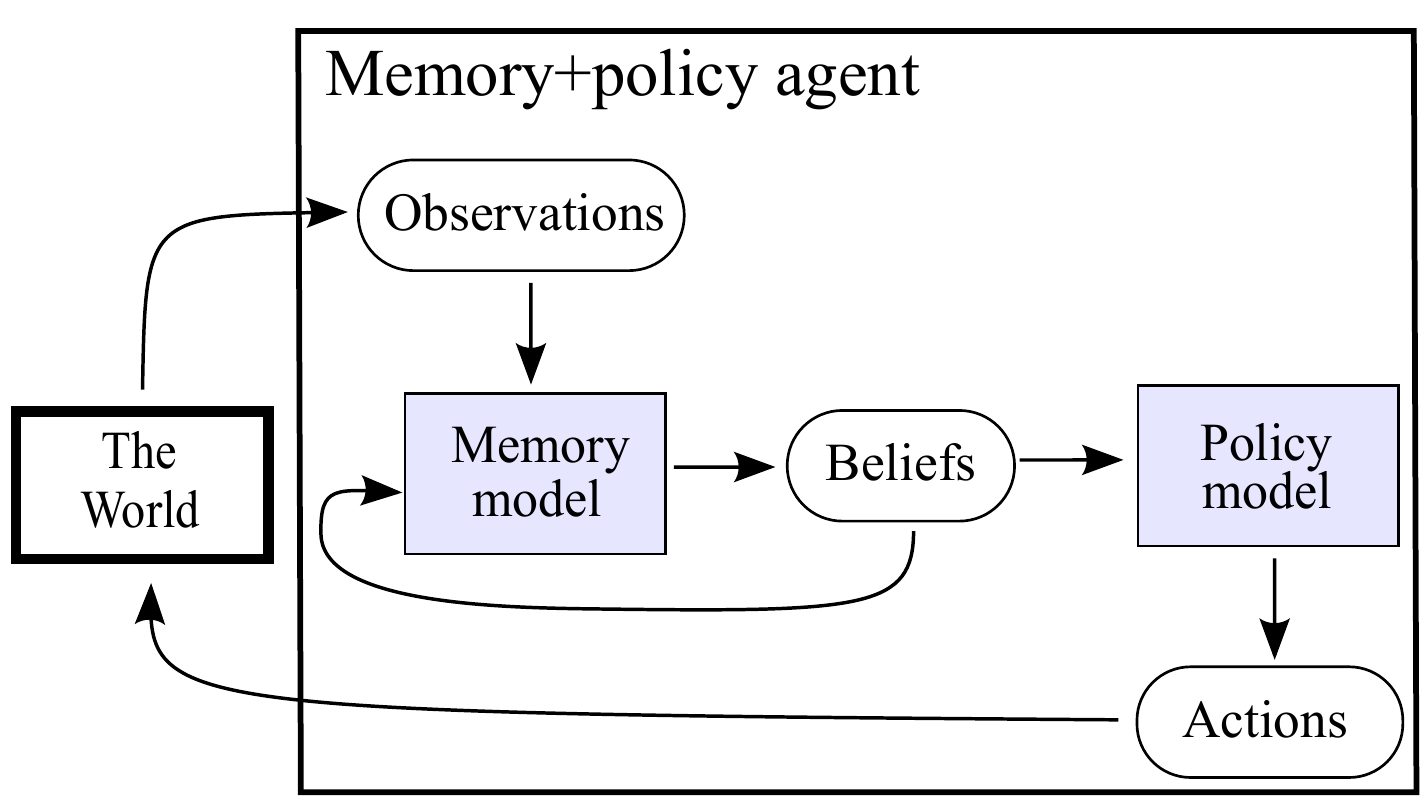}
		\caption{A simple architecture that is sufficient for general cognition,
			but is not practical to train.}
		\label{fig_memory_policy}
	\end{center}
\end{figure}

A diagram of a \emph{memory+policy agent} is given in Figure~\ref{fig_memory_policy}.
This architecture extends the policy agent with memory.
It uses two models, one to update its internal beliefs from new observations,
and one that maps from its current beliefs to actions.
This architecture can be shown to be theoretically sufficient for general cognition.
If its belief vector is sufficiently large, then it can represent any state that might occur in its world.
(This may not always be practical, but we are currently evaluating only the theoretical sufficiency of this
architecture.)
If its memory model is implemented with an arbitrary function approximator, then it can theoretically
update its beliefs as well as any other architecture could from the given observations.
If its policy model is implemented with an arbitrary function approximator,
then it can theoretically choose actions as well as any other architecture could from accurate beliefs.
Therefore, we can say that the memory+policy architecture is theoretically sufficient for general cognition.

Since the memory+policy model is theoretically sufficient for general cognition,
it would generally not be reasonable to resort to greater complexity for the purpose of
trying to achieve new theoretical capabilities.
However, the memory+policy architecture also has a significant practical limitation:
it is very difficult to train.
It requires a teacher that can unambiguously tell it which actions to choose in every
region of its belief space.
Even if a teacher is available to provide such thorough training, a memory+policy agent
would never be able to exceed the capabilities of its teacher, which renders it of
limited practical value.

The MANIC architecture adds some additional complexity so that it can separate
its learning system from its decision-making system.
This separation enables it to learn from unlabeled observations.
That is, it can refine the learning system's ability to accurately anticipate the consequences of actions
whether or not the teacher is available.
With MANIC, supervision is only needed to guide its priorities, not its beliefs or choices.
Yet, while being much more practical to train, MANIC is also provably sufficient for general cognition.

If we allow its belief vector, $\mathbf{v}_t$, to be arbitrarily large, then it it is sufficient
to encode a correct representation of the agent's world, no matter how complex that world may be.
If the transition function, $f$, is implemented with an arbitrary function approximator, then it is
theoretically sufficient to correctly anticipate any state transitions.
And, if the decoder, $g$, is also implemented with an arbitrary function approximator, then it will
be able to accurately anticipate observations from correct beliefs.
If we allow the genetic algorithm of the planning system to utilize an arbitrarily large population,
then it approximates an exhaustive search through all possible plans, which will find the optimal plan.
Since the action it chooses is always the first step in an optimal plan, its actions will be optimal
for maximizing contentment.
Finally, if the contentment model, $h$, is also implemented with a universal function approximator,
then it is sufficient to approximate the ideal utility metric for arbitrary problems.
Therefore, MANIC is sufficient for general cognition.
In the next Section, we discuss additional details about why MANIC is also practical for training.

%todo: If time in the agent's world is continuous, it can still approximate it to an arbitrary degree by sampling time at sufficiently-frequent discrete intervals. Therefore, we do not consider the discretization of time by MANIC to be limiting in any way.

%todo: It follows that the recurrent learning system consisting of $f$, $\mathbf{v}_t$, $g$, and $g^+$, is sufficient to implement a functional equivalence to human perception.

%todo: This reasoning does not prove that humans use similar mechanisms in their brains, but it does show that MANIC is theoretically sufficient to approximate whatever mechanisms humans use to learn from observations and plan their actions. Of course, other architectures can also achieve similar theoretical sufficiency, but lack clear methods for practical training. MANIC is significant because it achieves theoretical sufficiency while also being practical to train. In the next section, we describe how MANIC can be trained.

\section{Training}\label{sec_training}

Perhaps, the most significant insight for making MANIC practical for training comes from the high-level
division between the learning system and the decision-making system.
Because it divides where machine learning and artificial intelligence typically separate,
well-established training methods become applicable,
whereas such methods cannot be used with other cognitive architectures.
Because the learning system does not choose candidate actions, but only anticipates their effect,
it can learn from every observation that the agent makes, even if the actions are chosen randomly,
and even when no teacher is available.

The learning system in MANIC is a recurrent architecture because the current beliefs, $\mathbf{v}_t$, are used by the transition model to anticpate subsequent beliefs, $\mathbf{v}_{t+1}$.
Recurrent architectures are notoriously difficult to train, due to the chaotic effects of feedback that result from the recurrence
\cite{cuellar:train_recurrent_networks_difficult,sontag:neural_nets_for_control,sjoberg:nonlinearblackboxmodeling,floreano:evolve_recurrent_neural_nets,blanco:evolutionary_rnn}.
This tends to create both large hills and valleys throughout the error surface,
so local optimization methods must use very small learning rates. Further, local optimization methods are susceptible to the frequent local optima
that occur throughout the error space of recurrent architectures, and global optimization methods tend to be extremely slow.

However, recent advances in nonlinear dimensionality reduction provide a solution for cases where observations lie on a high-dimensional manifold.
Significantly, this situation occurs when observations consist of images that derive from continuous space.
In other words, MANIC can handle the case of a robot equipped with digital cameras, which has obvious analogy with humans equipped with optical vision.

\begin{figure}[!tb]
	\begin{center}
		\includegraphics[width=2.6in]{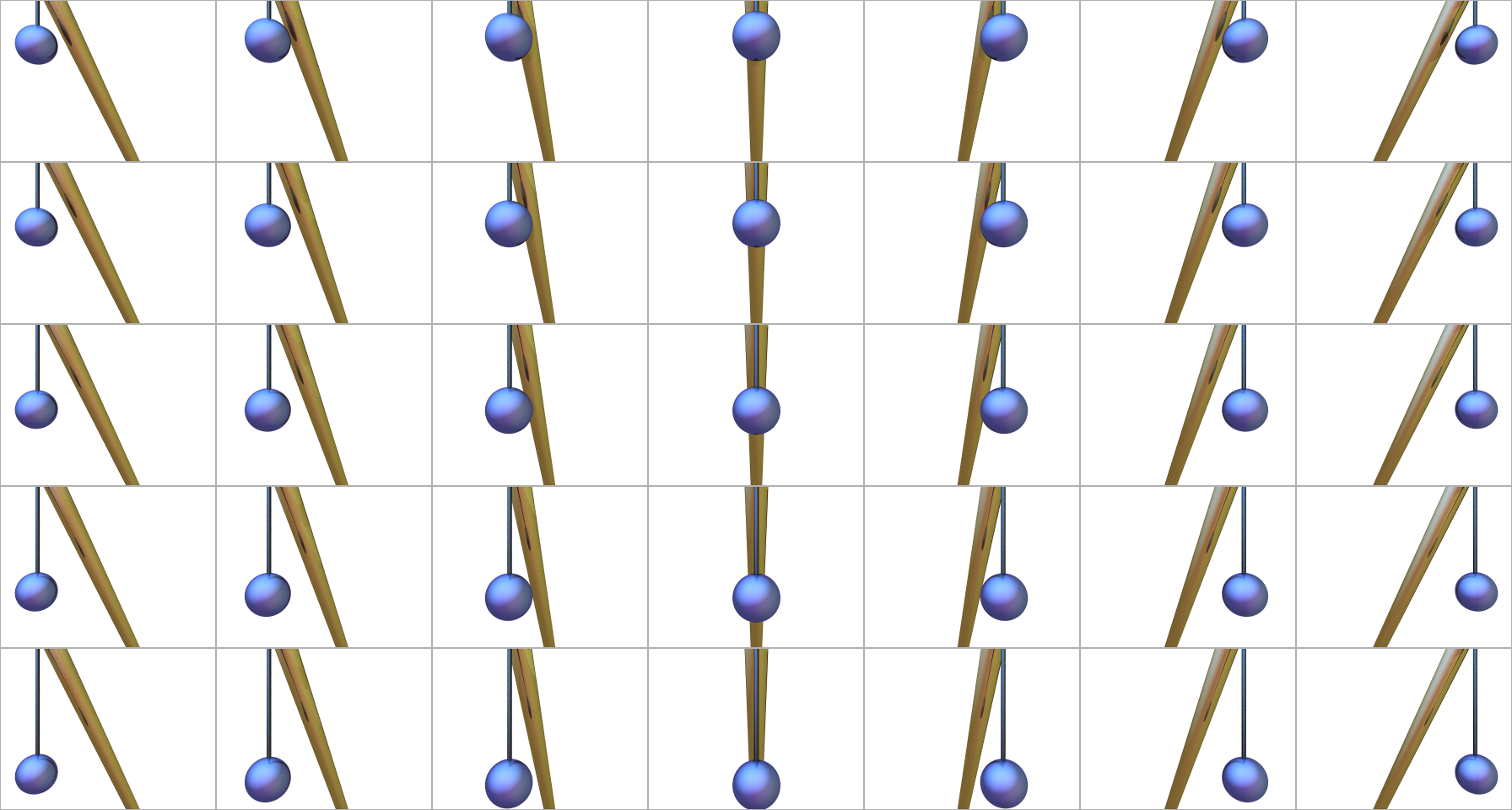}
		\caption{A system consisting of a simulated crane viewed through a camera. Each observation consists of
			9216 values, but the system itself only exhibits 2 degrees of freedom, so these images
			lie on a 2-dimensional non-linear manifold embedded in 9216-dimensional space. This figure
			depicts the entire manifold as represented by uniform sampling over its nonlinear surface.}
		\label{fig_crane}
	\end{center}
\end{figure}

\begin{figure}[!tb]
	\begin{center}
		\includegraphics[width=2.6in]{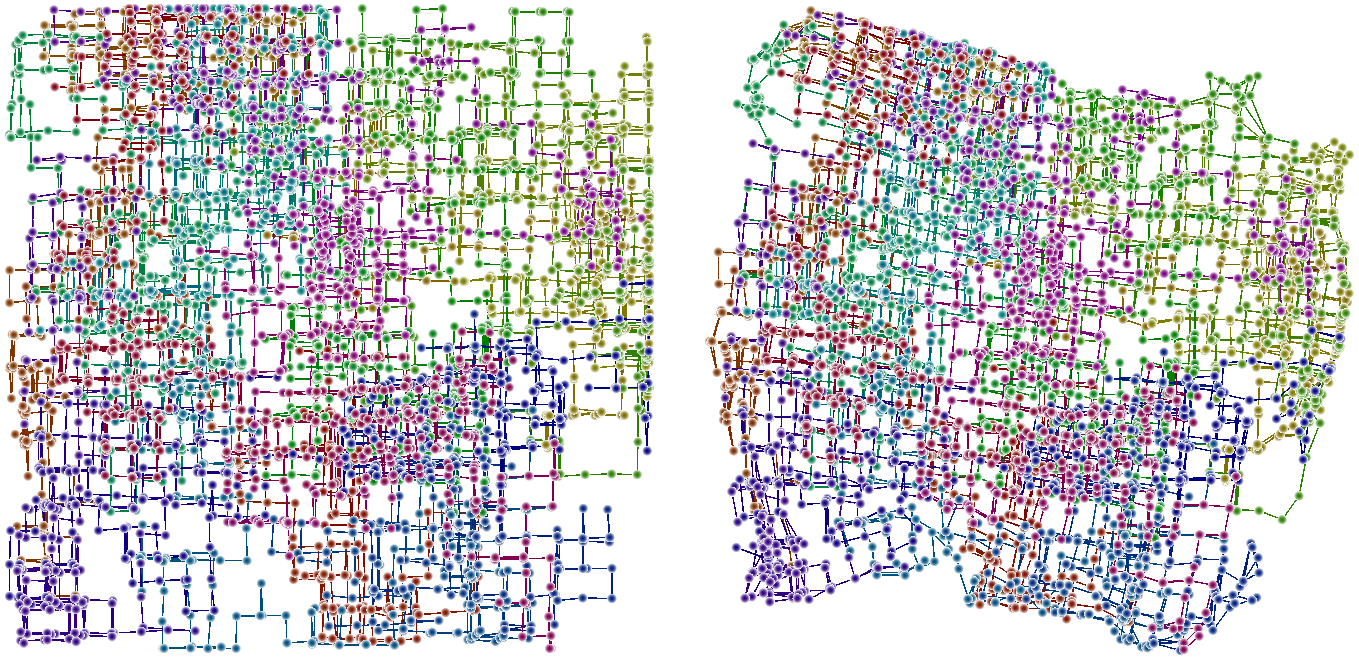}
		\caption{\textbf{Left:} The hidden states visited in a random walk with a simulated crane system.
			Color is used to depict time, starting with red and ending with purple.
			The horizontal axis shows boom position, and the vertical axis shows cable length. The model
			was shown images of the crane, but was not allowed to view the hidden state. \textbf{Right:}
			Estimated states calculated by reducing the dimensionality of observed images. Although differing
			somewhat from the actual states, these estimates were close enough to bootstrap training of
			a recurrent model of system dynamics.}
		\label{fig_crane_state}
	\end{center}
\end{figure}

When observations are high-dimensional, a good initial estimate of state (or in this case, beliefs) can be obtained by
reducing the dimensionality of those observations. %todo: cite something here
Co-author Gashler demonstrated this approach in 2011 with a method that trained deep recurrent artificial neural networks
to model dynamical systems \cite{gashler:tnldr}. For example, consider the crane system depicted in Figure~\ref{fig_crane}.
We used images containing $64\times48$ pixels in 3 color channels, for a total of 9216 values.
We performed a random walk through the state space of the crane by moving the crane left, right, up, or down,
at random to collect observations. (To demonstrate robustness, random noise was injected into state transitions
as well as the observed images.) Using a nonlinear dimensionality reduction technique, we reduced the 9216-dimensional
sequence of observed images down to just 2 dimensions (because the system has only 2 degrees of freedom) to obtain an estimate of the state represented in each
high-dimensional image. (See Figure~\ref{fig_crane_state}.) Significant similarity can be observed between the
actual (left) and estimated (right) states. Consequently, this approach is ideal for bootstrapping the training of a
recurrent model of system dynamics. When the beliefs are initialized to reasonable intial values, local optima
is much less of a problem, so regular stochastic gradient descent can be used to refine the model from subsequent
observations.

In the context of MANIC, this implies that nonlinear dimensionality reduction can be used to estimate each $\mathbf{v}_t$.
Then, $g$ can be trained to map from each $\mathbf{v}_t$ to $\mathbf{x}_t$,
$g^+$ can be trained to map from each $\mathbf{x}_t$ to $\mathbf{v}_t$, and $f$ can be trained to map from
each $\mathbf{v}_t$ to $\mathbf{v}_{t+1}$. Note that each of these mappings depends on having a reasonable estimate of $\mathbf{v}_t$.
These values are not typically known with recurrent architectures, but digital images provide sufficient
information that unsupervised dimensionality reduction methods can estimate $\mathbf{v}_1, \mathbf{v}_2, \cdots$
from $\mathbf{x}_1, \mathbf{x}_2, \cdots$ very well. When an estimate for each $\mathbf{v}_t$ is known, training the
various components of the learning system reduces to a simple supervised learning problem.

A similar three-step approach can be used to bootstrap the learning system of MANIC:

\begin{list}{$\bullet$}{}
	\item Gather high-dimensional observations.
	\item Reduce observations to an initial estimate of beliefs.
	\item Use beliefs to train the transition model.
\end{list}

The last step, training the transition model, is difficult with recurrent models because gradient-based
methods tend to get stuck in local optima. However, because dimensionality reduction can estimate
beliefs prior to training, it reduces to a simple supervised training approach. It has been shown that
bootstrapping the training of neural networks can effectively bypass the local optima that otherwise cause problems for refining
with gradient-based approaches \cite{bengio2007greedy}. This effect is much more dramatic with recurrent
neural networks, since they create local optima throughout their model space \cite{cuellar:train_recurrent_networks_difficult}.
Thus, after initial training, the system can be maintained using regular backpropagation to refine the
learning system from subsequent observations.

\begin{figure*}[!tb]
	\begin{center}
		\includegraphics[width=4.8in]{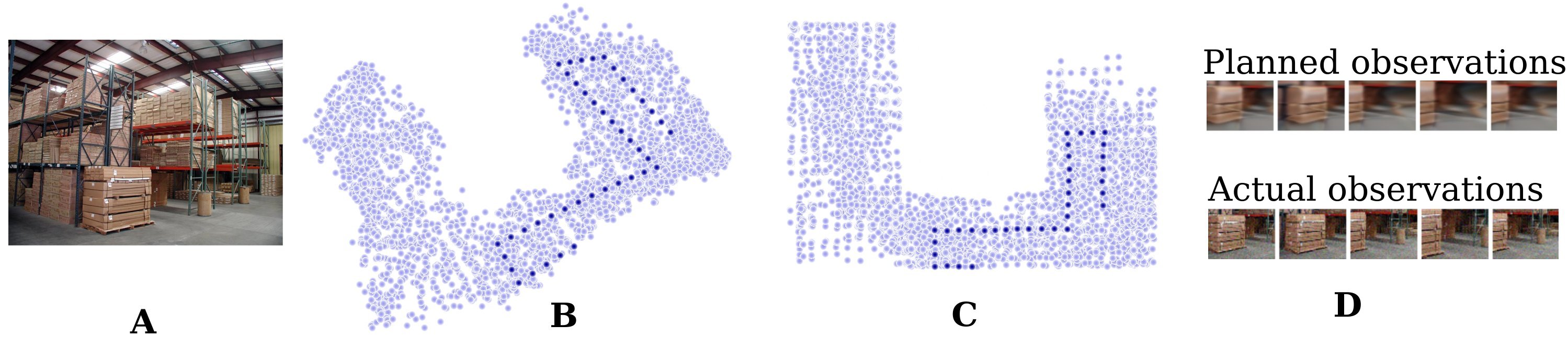}
		\caption{\textbf{A:} A model was trained to capture the dynamics of a robot that uses a camera
			to navigate within a warehouse. \textbf{B:} A sequence of actions was planned entirely within
			the belief-space of the model. \textbf{C:} When executed, the plan resulted in actual behavior
			corresponding with the agent's anticipated beliefs. \textbf{D:} A comparison of the agent's
			anticipated observations with those actually obtained when the plan was executed.}
		\label{fig_warehouse}
	\end{center}
\end{figure*}

With the crane dynamical system, we were able to accurately anticipate dynamics several hundred time-steps into the future \cite{gashler:tnldr},
even with injected noise.
To validate the plausibility of our planning system, we also demonstrated this approach on another problem involving a robot that navigates within a warehouse.
We first trained the learning system on a sequence of random observations. Then, using only observations predicted by the learning system,
we were able to successfully plan an entire sequence of actions by using the learning system to anticipate the ``beliefs" of MANIC.
The path we planned is shown in Figure~\ref{fig_warehouse}.B.
We, then, executed this plan on the actual system.
The actual states through which it passed are shown in Figure~\ref{fig_warehouse}.C.
Even though MANIC represented its internal belief states differently from the actual system, the anticipated observations were very close to the actual observations.

Many other recent advances in dimensionality reduction with deep artificial neural networks validate that this general approach is effective for
producing internal intrinsic representations of external observations.\cite{hinton2006reducing,vincent2008extracting,rifai2011contractive,bengio2011unsupervised,kingma2013auto}. %, also cite Ruslan Salakhutdinov's work, etc.]

The decision-making system contains one model, $h$, which needs to be trained to learn what constitutes homeostasis (or ``contentment") for the system.
This is done using a type of reinforcment learning. Because motivations are subjectively tied to human preferences,
the motivations for an agent that humans would receive as intellgent necessarily depends on human teaching.
Therefore, we assume that a human teacher is periodically available to direct the MANIC agent.
In cases where no human teacher is available, the contentment model could also be trained using a competitive or evolutionary approach.
This is done by instantiating multiple MANIC agents with variations in their contentment functions, and allowing the more fit
instantiations to survive.

When MANIC makes plans,
it utilizes its learning system to convert each plan from a sequence of actions
to a corresponding video of anticipated observations. In many ways, these videos of anticipated observations
may be analogous with the dreams or fantasies that humans produce internally as they sleep or anticipate future encounters.
Although differences certainly exist, this similar side-effect may indicate that the architecture within the human
brain has certain similarities with MANIC.
Ironically, the imagination of the artificial system is more accessible than that of biological humans,
enabling humans to examine the inner imaginings of the artificial system more intimately than they can
with each other.

The videos of imagined observations are presented to the human teacher (when he or she is available) for consideration.
The human, then, ranks these videos according to the disirability of the anticipated outcome.
Note that these plans need not actually be executed to generate the corresponding video. Rather, the teacher only ranks
imagined scenes. The precision with which the imagination of the agent corresponds with reality when it actually executes a
plan depends only on the learning system (which is continually refined), and does not depend on the teacher.
Because the teacher is only needed to refine the contentment model, only a reasonable amount of human attention is ever needed.

MANIC encapsulates a diversity of learning paradigms. The observation model, $g$, is trained
in an unsupervised manner from camera-based observations. The transition model, $f$, and the observation model, $g$ and $g^+$,
are trained by supervised learning. The contentment model, $h$, is trained by reinforcement from a human oracle. 
In 1999, Doya identified anatomical, physiological, and theoretical evidence to support the hypotheses that the cerebellum
specializes in supervised learning, the cerebral cortex specializes in unsupervised learning, and the basal ganglia specialize
in reinforcement learning \cite{doya1999computations}. This suggests a possible correlation between the components of MANIC
and those in the brain. Finally, the planning module in MANIC ties
all of its pieces together to make intelligent plans by means of a genetic algorithm. We consider it to be a positive property
that the MANIC architecture unifies such a wide diversity of learning techniques in such a natural manner. As each of these
methods exists for the purpose of addressing a particular aspect of cognition, we claim that a general cognitive architecture
must necessarily give place to each of the major paradigms that have found utility in artificial intelligence and machine learning.

\section{Sentience through self perception}\label{sec_sentience}

Sentience is a highly subjective and ill-defined concept, but is a critical aspect of the human experience.
Consequently, it has been the subject of much focus in cognitive architectures. % todo: need to cite something about this
We propose a plausible theory that might explain sentience, and show that MANIC can achieve functional equivalence.

It is clear that MANIC \emph{perceives} its environment, because it responds intelligently to the observations it makes.
Its perception is implemented by its beliefs, which describe its understanding of the observations it makes, as well as its observation model,
which connects its beliefs to the world.
The term ``awareness" is sometimes used with humans to imply a higher level of perception.
It is not clear whether MANIC achieves ``awareness", because the difference between ``awareness" and ``perception" is not yet well-defined.
However, because we have proven that MANIC is sufficient for general cognition, we can say for sure that MANIC achieves something
functionally equivalenct with awareness.
That is, we know it can behave as if it is aware, but we do not know whether human awareness requires certain immeasurable properties that MANIC lacks.

Similarly, the term ``sentience" contains aspects that are not yet well-defined, but other aspects of sentience are well-established.
Specifically, sentience requires that an agent possess feelings that summarize its overall well-being, as well as an awareness of those feelings,
and a desire to act on them.
An agent that implements the well-defined aspects of sentience can be said to implement something functionally equivalent.

Our definition of sentience is best expressed as an analogy with perception:
Perception requires the ability to observe the environment, beliefs that summarize or represent those observations, and a model to give the beliefs context.
Similarly, we propose that sentience requires the ability to make introspective observations, ``feelings" that summarize or represent them, and a model to give the feelings context.
In other words, if sentience arises from self-awareness, then MANIC can achieve something functionally equivalent through self perception.
In addition to a camera that observes the environment, MANIC can be equipped with the ability to observe its own internal state.
(For example, it might be enabled to additionally observe the weights of its three models, $f$, $g$, and $h$, and its belief vector, $\mathbf{v}$.)
Since MANIC is already designed for operating with high-dimensional observations, these introspective observations could simply be concatenated with the external observations it already makes.
This would cause MANIC to utilize a portion of $\mathbf{v}$ to summarize its introspective observations.

Thus, $\mathbf{v}$ would represent both the ``beliefs" and ``feelings" of a MANIC agent.
Its planning system would then implicilty make plans to maintaining homeostasis in both its beliefs and feelings.
And its observation  model would give context to its feelings by mapping between feelings and introspective observations, just as it does between beliefs and external observations.
This theory of sentience is plausible with humans,
because humans plan with regard to both their feelings and beliefs as a unified concept,
maintaining both their external objectives and internal well-being,
and because a well-connected brain, which humans have, would be sufficient to provide the ``introspective observations" necessary to facilitate it.

Since MANIC learns to model its priorities from a human teacher in its contentment function, $h$,
it will learn to give appropriate regard to its own feelings when they are relevant for its purpose.
Presumably, this will occur when the human teacher directs it to maintain itself.
Using the same observation model with both introspective and external observations,
and using the same vector to model both feelings and beliefs,
are both plausible because these design choices will enable MANIC to entangle its feelings with its environment,
behavior that humans are known to exhibit.

\section{Conclusion}\label{sec_conclusion}

We presented a cognitive architecture called MANIC. This architecture unifies a diversity of techniques in
the sub-disciplines of machine learning and artificial intelligence without introducing much novelty.
Yet, while relying on existing methods, and with minimal complexity, MANIC is a powerful cognitive architecture.
We showed that it is sufficiently general to accomplish arbitrary cognitive tasks, and that it can be practically
trained using recent methods.

We supported MANIC's design by referring to existing works
that validate its individual components, and we made theoretical arguments about the capabilities that should
emerge from combining them in the manner outlined by MANIC. The primary contribution of this paper is to show that
these existing methods can already accomplish more of cognitive intelligence than is generally recognized.
Our ultimate intent is to argue that if general intelligence is one of the ultimate objectives
of the fields of machine learning and artificial intelligence, then they are very much on the right track,
and it is not clear that any critical piece of understanding necessary for implementing a rudimentary consciousness
is definitely missing.

\bibliographystyle{IEEEtran}
\bibliography{refs}

\end{document}